\pdfoutput=1
\documentclass[14pt]{article}

\usepackage{arxiv}
\usepackage{times}
\usepackage{longtable}
\usepackage{algorithm}
\usepackage{algorithmic}
\usepackage{nicefrac}
\usepackage{booktabs}
\usepackage{footmisc}
\usepackage[shortlabels]{enumitem}
\usepackage{multirow}

\usepackage{array}
\usepackage{url}
\usepackage[strict]{changepage}
\usepackage{makecell}
\usepackage{titlesec}
\usepackage{setspace}
\usepackage{bm}
\usepackage{mdframed}
\usepackage{bbm}
\usepackage{threeparttable}
\usepackage{subfigure}

\usepackage[T1]{fontenc}

\usepackage{titletoc}
\usepackage{longtable}
\usepackage{pifont} 
\usepackage{wrapfig}

\usepackage[table]{xcolor}
\usepackage{makecell}

\usepackage{fancyhdr} 

\usepackage[most]{tcolorbox} 

\definecolor{darkpastelgreen}{rgb}{0.01, 0.75, 0.24}

\titleformat*{\subparagraph}{\itshape}

\newsavebox\actorsfigure

\usepackage{circledsteps}
\usepackage{fontawesome5}

\definecolor{LightGray}{RGB}{250,250,250}
\definecolor{Gray}{RGB}{240, 240, 240}

\definecolor{quoteLine}{RGB}{0,0,0} 
\definecolor{quoteText}{RGB}{0,0,0} 
\definecolor{authorText}{RGB}{0,0,0} 
\definecolor{hidden-draw}{RGB}{106,142,189} 
\definecolor{hidden-blue}{RGB}{194,230,247} 
\definecolor{hidden-orange}{RGB}{217, 230, 252}
\definecolor{hidden-red}{RGB}{189,106,118}    
\definecolor{hidden-yellow}{RGB}{189,175,106} 
\definecolor{hidden-green}{RGB}{106,189,142}  
\definecolor{hidden-orange}{RGB}{189,142,106} 
\definecolor{hidden-purple}{RGB}{142,106,189} 

\newcommand{\litmark}{\textcolor{hidden-draw}{\ding{108}}}     
\newcommand{\ideamark}{\textcolor{hidden-orange}{\ding{108}}}   
\newcommand{\expmark}{\textcolor{hidden-red}{\ding{108}}}     
\newcommand{\execmark}{\textcolor{hidden-yellow}{\ding{108}}}  
\newcommand{\writmark}{\textcolor{hidden-green}{\ding{108}}}    
\newcommand{\papermark}{\textcolor{hidden-purple}{\ding{108}}}   

\NewDocumentCommand{\heng}
{ mO{} }{\textcolor{red}{\textsuperscript{\textit{Heng}}\textsf{\textbf{\small[#1]}}}}
\usepackage[numbers]{natbib}
\usepackage{amsfonts}
\usepackage[hidelinks]{hyperref}

\title{A Survey of AI Scientists}

\author{
{\bfseries Guiyao Tie$^{1}$\footnotemark[2]}\quad
{\bfseries Pan Zhou$^{1}$}\quad
{\bfseries Lichao Sun$^{2}$}\quad
\\
\\
{\bfseries $^{1}$Huazhong University of Science and Technology}\quad
{\bfseries $^{2}$Lehigh University}\quad\\
}

\begin{document}
\maketitle
\renewcommand{\thefootnote}{\fnsymbol{footnote}}

\footnotetext[2]{Guiyao Tie is the current corresponding author: \href{mailto:tgy@hust.edu.cn}{tgy@hust.edu.cn}}
\footnotetext[3]{Latest Update: Oct., 2025.}


\begin{abstract}
A new scientific paradigm, the AI Scientist, has coalesced at the intersection of artificial intelligence and epistemology, promising a fundamental shift from AI-assisted analysis to end-to-end autonomous discovery. Catalyzed by rapid advances in large language models, multi-agent orchestration, and robotic automation, these systems are architected to emulate the complete scientific workflow—from initial hypothesis generation to the final synthesis of publishable findings. This transition moves beyond using AI as an instrument of inquiry, positioning it as a potential originator of scientific knowledge. However, the rapid and unstructured proliferation of these systems has created a fragmented research landscape, obscuring overarching methodological principles and developmental trends. This survey provides a systematic and comprehensive synthesis of this emerging domain by introducing a unified, six-stage methodological framework that deconstructs the scientific process into: Literature Review, Idea Generation, Experimental Preparation, Experimental Execution, Scientific Writing, and Paper Generation. 
Through this analytical lens, we systematically map and analyze dozens of seminal works from 2022 to late 2025, revealing a clear three-phase evolutionary trajectory. Our analysis charts the field's progression from an initial phase of Foundational Modules, focused on task-specific automation, through a period of Closed-Loop Integration, to the current frontier of Scalability, Impact, and Collaboration. By synthesizing these developments, this survey identifies key architectural patterns and highlights the dual research thrusts toward both greater machine autonomy and more sophisticated human-in-the-loop synergy. We conclude by presenting a forward-looking agenda that addresses critical open challenges in robustness, generalizability, and ethical governance. Ultimately, this work provides a critical roadmap for the field, intended to guide the next generation of systems toward becoming trustworthy, verifiable, and indispensable partners in human scientific inquiry.
Project Github: \textcolor{red}{ \underline{\href{https://github.com/Mr-Tieguigui/Survey-for-AI-Scientist}{https://github.com/Mr-Tieguigui/Survey-for-AI-Scientist}}.}
\end{abstract}





\newpage
\tableofcontents
\newpage
\section{Introduction}\label{Section 1}

Artificial intelligence (AI) is undergoing a fundamental transformation in its role within scientific research. Historically deployed as a computational instrument for specific analytical tasks (such as pattern recognition, data mining, and predictive modeling~\citep{liu2022real, almarie2023editorial, erduran2024impact}), AI is now emerging as an autonomous originator of scientific knowledge. This shift represents more than incremental technological progress; it embodies a qualitative leap from \emph{AI for Science}, where AI augments human-defined workflows, to \emph{AI as Scientist}, where AI systems independently conceptualize, execute, and communicate original research. The \textbf{AI scientist} paradigm pursues end-to-end autonomous discovery—executing the complete scientific workflow from hypothesis formulation through experimental execution to manuscript generation~\citep{gridach2025agentic, wei2025ai, Yamada2025AIScientistV2}. This transformation is enabled by three converging technological advances: (1) large language models with emergent reasoning capabilities~\citep{OpenAI2024GPT4}; (2) multi-agent architectures enabling complex task decomposition and collaboration~\citep{schmidt2009distilling, Ghafarollahi2024SciAgents}; and (3) robotic experimentation platforms bridging digital reasoning with physical execution~\citep{Curie2025, BioPlanner2023}. Collectively, these advances mark a transition from automation—wherein AI supports predefined protocols—to autonomy—wherein AI independently designs, adapts, and validates investigative processes.

The intellectual lineage of autonomous scientific discovery extends to early symbolic AI systems~\citep{Langley1987ScientificDiscovery, schmidt2009distilling}, which automated rediscovery of established physical laws through symbolic regression and rule-based reasoning. However, these pioneering efforts were constrained by brittle knowledge representations, limited generalization beyond carefully curated domains, and inability to interface with real-world experimentation. Contemporary AI Scientist systems transcend these limitations by integrating foundation models with closed-loop scientific reasoning (\textbf{observe → hypothesize → experiment → analyze → publish}), through agentic planning, principle-driven inference, and adaptive self-correction mechanisms. This integration enables systems to operate across heterogeneous knowledge modalities, navigate ambiguous problem spaces, and iteratively refine hypotheses based on empirical feedback. Representative implementations~\citep{Lu2024AIScientist, Curie2025, PiFlow2025} demonstrate cross-domain capabilities spanning chemical synthesis~\citep{Boiko2023Coscientist}, genomic network inference~\citep{Afonja2024LLM4GRN}, and symbolic equation discovery~\citep{SRScientist2025}, validating the generality of the autonomous research paradigm.

\begin{figure}[ht]
\centering
\includegraphics[width=1\linewidth]{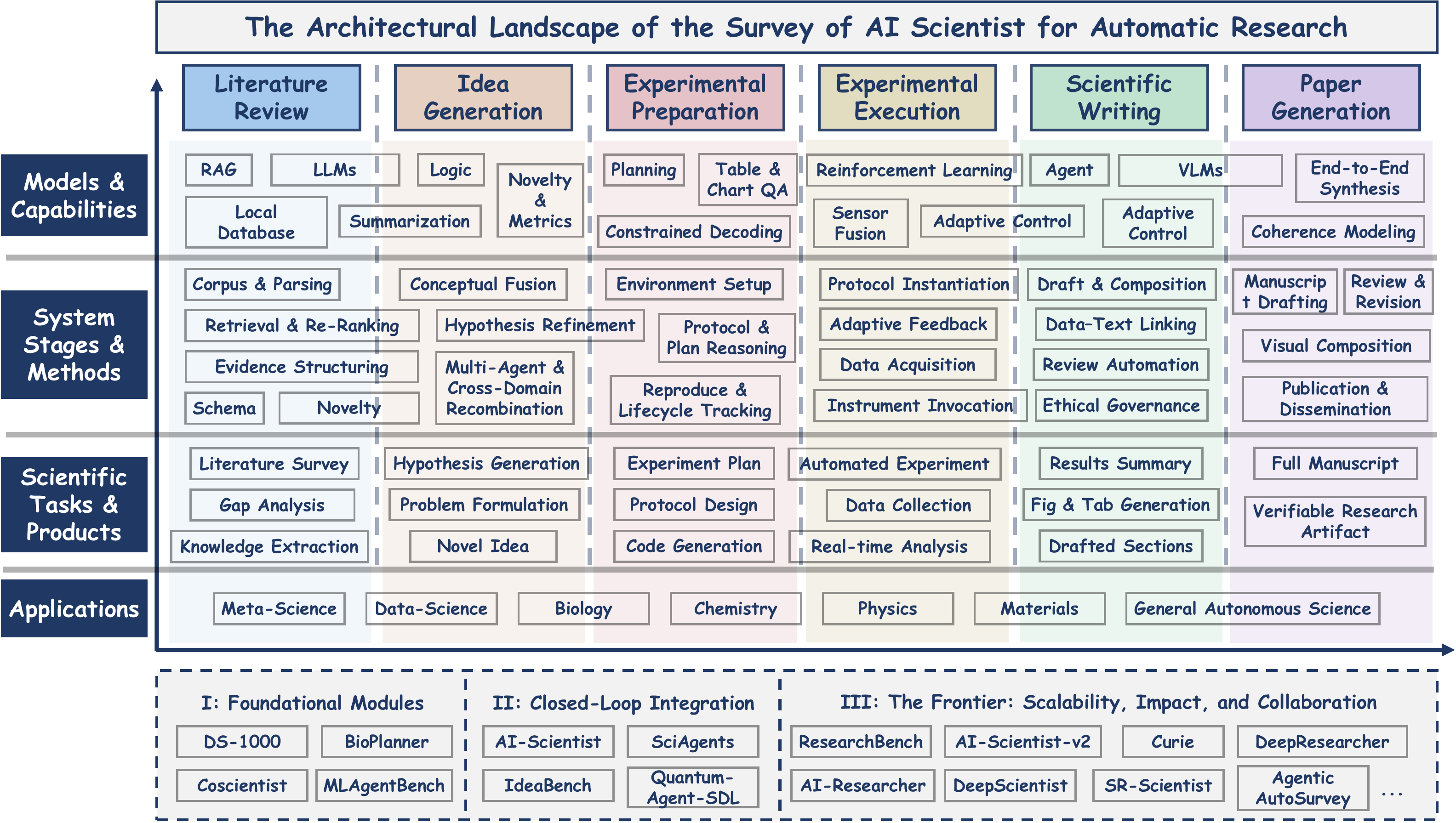}
\caption{
\textbf{The Architectural Landscape of the AI Scientists for Automatic Research.}
The main 4x6 matrix maps the six methodological stages of the scientific workflow (horizontal axis) against four top-down layers of abstraction, from Applications to Models (vertical axis). The panel at the bottom illustrates the field's three-phase historical evolution, categorizing representative works to provide a chronological perspective on the development of AI Scientist systems.
}
\label{fig:organization}
\end{figure}

Despite these advances, the field faces a critical challenge: \emph{conceptual fragmentation}. The rapid proliferation of AI Scientist systems—evidenced by specialized frameworks achieving near-human performance~\citep{Yamada2025AIScientistV2, Tang2025AIResearcher} and domain-specific benchmarks evaluating novelty, causality, and reproducibility~\citep{chen2025auto, liu2025researchbench, guo2025ideabench}—has outpaced the development of unifying theoretical frameworks. Current research remains fragmented across three dimensions: (1) \emph{methodological silos}, where systems excel at isolated stages (e.g., hypothesis generation~\citep{Zhou2024HypothesisGenerationLLMs, vasu2025hyper}, literature synthesis~\citep{Agarwal2024LitLLM}, or experimental execution~\citep{Curie2025}) without addressing end-to-end integration; (2) \emph{domain-specific architectures}, optimized for chemistry~\citep{Boiko2023Coscientist}, biology~\citep{BioPlanner2023}, or physics with limited cross-domain generalization; and (3) \emph{evaluation heterogeneity}, where benchmarks assess non-overlapping capabilities without establishing unified quality metrics. While existing surveys~\citep{gridach2025agentic, wei2025ai, luo2025llm4sr} provide valuable domain-specific or capability-focused reviews, no comprehensive synthesis systematically connects methodological stages, architectural patterns, and evaluation protocols across the full spectrum of AI Scientist research. This conceptual gap impedes cumulative progress, obscures transferable insights, and hinders the identification of fundamental research challenges versus domain-specific engineering issues.

This survey synthesizes research from 2022–2025 through a unified architectural framework (\textbf{Figure~\ref{fig:organization}}). The horizontal axis defines six methodological stages: (1)~Literature Review: knowledge structuring~\citep{Agarwal2024LitLLM}; (2)~Idea Generation: hypothesis formulation~\citep{vasu2025hyper}; (3)~Experimental Preparation: protocol design~\citep{Lai2023DS1000, guo2024ds}; (4)~Experimental Execution: adaptive orchestration~\citep{BioPlanner2023, Curie2025, Panapitiya2025AutoLabs}; (5)~Scientific Writing: multimodal composition~\citep{WritingBench2025Benchmark}; and (6)~Paper Generation: manuscript synthesis~\citep{Tang2025AIResearcher, Lu2024AIScientist}. The vertical axis stratifies four abstraction layers—Domains, Tasks, Architectures, Capabilities—forming a 4×6 categorization matrix. The timeline panel traces three evolutionary phases: Foundational Modules, Closed-Loop Integration, and Scalable Collaboration.

\subsection{Contributions}
This survey provides a systematic synthesis of AI Scientist research through three primary contributions:

\begin{itemize}[leftmargin=1.3em]
    \item \textbf{Unified Six-Stage Framework.} We formalize the autonomous research pipeline through six interconnected stages—Literature Review, Idea Generation, Experimental Preparation, Experimental Execution, Scientific Writing, and Paper Generation—modeling information dependencies and stage-specific capabilities. This framework reveals transferable methodological principles across scientific disciplines.
    
    \item \textbf{Comprehensive Systematic Categorization.} We analyze 50+ representative systems spanning 2022–2025, mapping each to our six-stage framework to reveal methodological patterns and capability distributions. This categorization shows that most systems excel at 2-3 stages while remaining nascent in others, and that domain-specific systems converge on similar stage-level architectures.
    
    \item \textbf{Three-Phase Historical Analysis.} We identify a coherent developmental progression: Foundational Modules (2022–2023) establishing stage-specific capabilities; Closed-Loop Integration (2024) unifying multiple stages; and Scalable Collaboration (2025–present) pursuing robustness and human-AI partnership. Systems like The AI Scientist exemplify the acceleration from isolated capabilities to full paper generation within 18 months.
\end{itemize}

\subsection{Organization}
\textbf{Section~\ref{sec:background}} presents the taxonomy and historical evolution.  
\textbf{Section~\ref{sec:method}} details methodological advances across the six stages.  
\textbf{Section~\ref{sec:applications}} examines domain-specific implementations.  
\textbf{Section~\ref{sec:problem}} identifies open challenges and future directions.

\section{Background and Taxonomy}
\label{sec:background}

This chapter establishes the conceptual framework for our survey by introducing a taxonomy of research categories and tracing the historical evolution of AI scientist systems. We begin by deconstructing the end-to-end scientific process into six distinct methodological stages. This classification serves as the primary analytical lens through which we categorize existing works. We then present a comprehensive matrix aligning representative systems and benchmarks from 2022 to 2025 with these stages, built upon a rigorous, verifiable review of each cited paper. Finally, we synthesize these findings into a historical narrative, identifying three major phases of development that chart the field's progression from modular tools to integrated, self-reflective research agents.

\begin{table*}[t]
\centering
\renewcommand{\arraystretch}{1.15}
\caption{\textbf{Comprehensive matrix of AI Scientist works (updated to 2025) aligned with six methodological stages.} This table has been rebuilt and expanded based on a rigorous, verifiable review of each paper's primary contributions. Each colored symbol represents explicit coverage of a methodological stage.}
\label{tab:category_matrix}
\footnotesize
\begin{tabular}{
>{\raggedright\arraybackslash}p{3.5cm}
>{\centering\arraybackslash}p{0.7cm}
>{\centering\arraybackslash}p{0.7cm}
>{\centering\arraybackslash}p{0.7cm}
>{\centering\arraybackslash}p{0.7cm}
>{\centering\arraybackslash}p{0.7cm}
>{\centering\arraybackslash}p{0.7cm}
>{\centering\arraybackslash}p{1.3cm}}
\toprule
\textbf{Work} & \rotatebox{0}{\textbf{\textcolor{hidden-draw}{Lit.}}} & \rotatebox{0}{\textbf{\textcolor{hidden-orange}{Idea}}} & \rotatebox{0}{\textbf{\textcolor{hidden-red}{Exp.}}} & \rotatebox{0}{\textbf{\textcolor{hidden-yellow}{Exec.}}} & \rotatebox{0}{\textbf{\textcolor{hidden-green}{Writ.}}} & \rotatebox{0}{\textbf{\textcolor{hidden-purple}{Paper}}} & \textbf{Year} \\
\midrule
DS-1000~\citep{Lai2023DS1000} & & & \expmark & \execmark & & & 2023.04 \\
Coscientist~\citep{Boiko2023Coscientist} & & \ideamark & \expmark & \execmark & & & 2023.06 \\
BioPlanner~\citep{BioPlanner2023} & & & \expmark & & & & 2023.10 \\
MLAgentBench~\citep{huang2023mlagentbench} & & & \expmark & \execmark & & & 2023.10 \\
\midrule
LitLLM~\citep{Agarwal2024LitLLM} & \litmark & & & & \writmark & & 2024.02 \\
AI Scientist (v1)~\citep{Lu2024AIScientist} & \litmark & \ideamark & \expmark & \execmark & \writmark & \papermark & 2024.08 \\
SciAgents~\citep{Ghafarollahi2024SciAgents} & \litmark & \ideamark & & & & & 2024.09 \\
IdeaBench~\citep{guo2025ideabench} & & \ideamark & & & & & 2024.11 \\
Quantum-Agent-SDL~\citep{Cao2024QuantumAgentSDL} & & \ideamark & \expmark & \execmark & & & 2024.12 \\
\midrule
HypER~\citep{vasu2025hyper} & \litmark & \ideamark & & & & & 2025.01 \\
AI Scientist (v2)~\citep{Yamada2025AIScientistV2} & \litmark & \ideamark & \expmark & \execmark & \writmark & \papermark & 2025.02 \\
Curie~\citep{Curie2025} & & \ideamark & \expmark & \execmark & & & 2025.02 \\
AI co-scientist~\citep{gottweis2025towards} & \litmark & \ideamark &  &  & \writmark & & 2025.02 \\
ResearchBench~\citep{liu2025researchbench} & \litmark & \ideamark & \expmark & \execmark & \writmark & & 2025.03 \\
DeepResearcher~\citep{Zheng2025DeepResearcher} & \litmark & \ideamark & \expmark & \execmark & \writmark & & 2025.04 \\
AutoLabs~\citep{Panapitiya2025AutoLabs} & & \ideamark & \expmark & \execmark & & & 2025.04 \\
AI-Researcher~\citep{Tang2025AIResearcher} & \litmark & \ideamark & \expmark & \execmark & \writmark & \papermark & 2025.05 \\
EXP-Bench~\citep{Kon2025EXPBench} & & \ideamark & \expmark & \execmark & \writmark & & 2025.05 \\
Agentic AutoSurvey~\citep{liu2025agentic} & \litmark & \ideamark &  &  & \writmark & & 2025.09 \\
PiFlow~\citep{PiFlow2025} & \litmark & \ideamark & \expmark & \execmark & & & 2025.09 \\
DeepScientist~\citep{Weng2025DeepScientist} & \litmark & \ideamark & \expmark & \execmark & \writmark & \papermark & 2025.09 \\
SR-Scientist~\citep{SRScientist2025} & & \ideamark & \expmark & \execmark & & & 2025.10 \\
Freephdlabor~\citep{Li2025FreephdLabor} & \litmark & \ideamark & \expmark & \execmark & \writmark & \papermark & 2025.10 \\
\bottomrule
\end{tabular}
\end{table*}

\subsection{Taxonomy of Research Categories}
\label{sec:taxonomy}

AI Scientist research from 2022 to 2025 can be systematically deconstructed into six \textbf{methodological stages}. Each stage represents a critical phase of the scientific process that has been progressively automated. Together, they form an end-to-end pipeline that transforms unstructured knowledge into verifiable scientific output, bridging abstract cognition with concrete execution and communication.

\begin{itemize}[leftmargin=1.6em]
    \item[\textcolor{hidden-draw}{$\bullet$}] \textbf{\textcolor{hidden-draw}{\underline{Literature Review (Lit.).}}} This foundational stage involves transforming unstructured scientific corpora into machine-interpretable knowledge. It encompasses techniques from large-scale information retrieval to the synthesis of research gaps. Systems like LitLLM~\citep{Agarwal2024LitLLM} and HypER~\citep{vasu2025hyper} focus on citation-grounded summarization and knowledge extraction, while advanced frameworks like SciAgents~\citep{Ghafarollahi2024SciAgents} and DeepResearcher~\citep{Zheng2025DeepResearcher} utilize web-scale interactions and graph reasoning to map existing knowledge and provide a robust foundation for downstream tasks.

    \item[\textcolor{hidden-orange}{$\bullet$}] \textbf{\textcolor{hidden-orange}{\underline{Idea Generation (Idea).}}} Building upon the structured knowledge from the prior stage, this phase automates hypothesis discovery and problem formulation. It leverages the creative and reasoning capabilities of LLMs to propose novel yet plausible research directions. This capability is explicitly evaluated by benchmarks like IdeaBench~\citep{guo2025ideabench}, and is a core component of both domain-specific systems like Coscientist~\citep{Boiko2023Coscientist} and advanced end-to-end frameworks like DeepScientist~\citep{Weng2025DeepScientist}.

     \item[\textcolor{hidden-red}{$\bullet$}] \textbf{\textcolor{hidden-red}{\underline{Experimental Preparation (Exp.).}}} This crucial intermediate stage translates an abstract hypothesis into an executable plan. It includes tasks such as defining variables, selecting datasets, generating analysis code, and designing experimental protocols. This is a primary focus of data-science-oriented benchmarks like DS-1000~\citep{Lai2023DS1000} and MLAgentBench~\citep{huang2023mlagentbench}, and is a key step in all integrated systems from The AI Scientist v1~\citep{Lu2024AIScientist} to freephdlabor~\citep{Li2025FreephdLabor}.

     \item[\textcolor{hidden-yellow}{$\bullet$}] \textbf{\textcolor{hidden-yellow}{\underline{Experimental Execution (Exec.).}}} This stage involves the actual running of real or simulated experiments. It emphasizes the agent's ability to interact with tools, control robotics, and adapt its plan based on real-time feedback. Milestones in this area include systems that orchestrate physical laboratory hardware, such as Coscientist~\citep{Boiko2023Coscientist} and Quantum-Agent-SDL~\citep{Cao2024QuantumAgentSDL}. Frameworks like Curie~\citep{Curie2025} and DeepResearcher~\citep{Zheng2025DeepResearcher} demonstrate this capability in simulated and real-world web environments, respectively.

     \item[\textcolor{hidden-green}{$\bullet$}] \textbf{\textcolor{hidden-green}{\underline{Scientific Writing (Writ.).}}} This stage focuses on the communication of scientific findings by transforming structured results into coherent, citation-grounded narratives. Capabilities range from section-aware summarization to data-to-text synthesis. This is a key feature in end-to-end systems like ResearchBench~\citep{liu2025researchbench} and is central to human-in-the-loop frameworks like freephdlabor~\citep{Li2025FreephdLabor}, where the AI drafts content for human review and refinement.

     \item[\textcolor{hidden-purple}{$\bullet$}] \textbf{\textcolor{hidden-purple}{\underline{Paper Generation (Paper).}}} Representing the culmination of the scientific workflow, this final stage synthesizes a full, publication-ready manuscript. This requires the tight integration of all prior stages. This end-to-end capability is the hallmark of the most advanced, fully autonomous systems, such as The AI Scientist v1/v2~\citep{Lu2024AIScientist, Yamada2025AIScientistV2}, AI-Researcher~\citep{Tang2025AIResearcher}, and DeepScientist~\citep{Weng2025DeepScientist}.
\end{itemize}

\begin{figure}[t]
\centering
\includegraphics[width=1\linewidth]{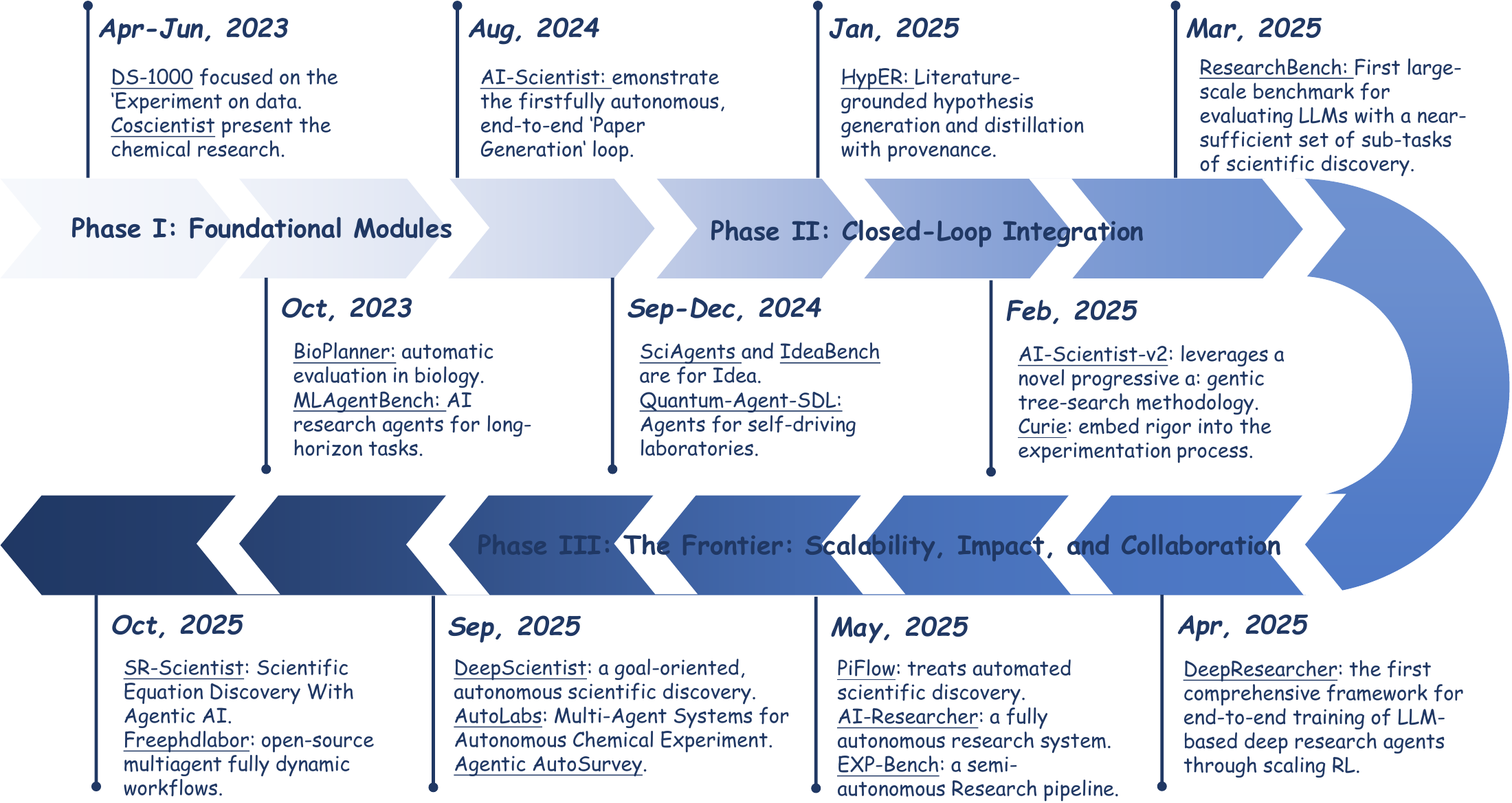}
\caption{
\textbf{Evolution of AI Scientist research (2022–2025).} A horizontal timeline illustrates three major phases: \emph{(I) Foundational Modules (2022–2023)}, \emph{(II) Closed-Loop Integration (2024)}, and \emph{(III) The Frontier: Scalability, Impact, and Collaboration (2025–present)}. Each phase highlights representative systems, with upward arrows denoting increasing levels of autonomy and integration.
}
\label{fig:history}
\end{figure}

\subsection{Historical Evolution}
\label{sec:history}

The evolution of AI Scientist research from 2022 to 2025 reveals a clear trajectory: a progressive integration of automation, moving from discrete, task-specific modules toward self-reflective and verifiable end-to-end systems. We identify three major developmental phases that capture this conceptual deepening.

\begin{itemize}
    \item \textbf{Phase I: Foundational Modules (2022–2023).} This initial phase established the foundational infrastructure for automating individual scientific workflow stages. Experimental preparation and execution received primary attention through benchmarks like DS-1000~\citep{Lai2023DS1000}, which formalized data science code generation tasks, and MLAgentBench~\citep{huang2023mlagentbench}, which evaluated autonomous experimentation in machine learning contexts. Protocol design capabilities emerged through BioPlanner~\citep{BioPlanner2023}, which translated natural language into executable biological protocols, establishing standards for biological experimental automation. Physical laboratory automation achieved a milestone through Coscientist~\citep{Boiko2023Coscientist}, which integrated LLMs with robotic liquid handlers to demonstrate closed-loop chemical synthesis, validating the feasibility of AI-driven physical experimentation. Concurrently, early work in quantum computing experimentation~\citep{Cao2024QuantumAgentSDL} explored autonomous calibration and error correction, extending the automation paradigm to frontier physics domains. These systems collectively established modular capabilities across isolated workflow stages while revealing integration challenges.
    \item \textbf{Phase II: Closed-Loop Integration (2024).} The year 2024 witnessed a paradigm shift from isolated modules toward integrated workflows spanning multiple scientific stages. Literature synthesis capabilities matured through LitLLM~\citep{Agarwal2024LitLLM}, which demonstrated citation-grounded knowledge extraction and narrative generation, establishing foundations for autonomous literature review. Hypothesis generation emerged as a distinct research thrust through SciAgents~\citep{Ghafarollahi2024SciAgents}, which employed multi-agent collaboration for scientific ideation, and IdeaBench~\citep{guo2025ideabench}, which formalized evaluation metrics for research question quality, novelty, and feasibility. The defining achievement of this phase was The AI Scientist v1~\citep{Lu2024AIScientist}, which unified literature review, idea generation, experimental execution, and manuscript writing into the first fully autonomous paper generation loop. This system demonstrated end-to-end research automation in machine learning domains, generating complete papers with minimal human intervention. Concurrently, specialized systems extended closed-loop reasoning to quantum computing~\citep{Cao2024QuantumAgentSDL} and materials science~\citep{Szymanski2023AutonomousLab}, validating cross-domain generalization of integrated workflows.
    \item \textbf{Phase III: Scalability, Impact, and Collaboration (2025–present).} The contemporary frontier pursues three convergent objectives: scalability, scientific impact, and human-AI collaboration. \emph{Scalability and robustness} advances through systems employing reinforcement learning and adaptive planning. HypER~\citep{vasu2025hyper} introduces provenance-aware hypothesis refinement through knowledge graph reasoning, while The AI Scientist v2~\citep{Yamada2025AIScientistV2} implements agentic tree-search architectures enabling parallel exploration of research directions. DeepResearcher~\citep{Zheng2025DeepResearcher} extends scalability through reinforcement learning in web environments, training agents to navigate noisy, unstructured information. PiFlow~\citep{PiFlow2025} demonstrates principled scientific reasoning across domains through modular workflow orchestration. \emph{Scientific impact} is targeted through systems designed for frontier discovery. ResearchBench~\citep{liu2025researchbench} establishes comprehensive evaluation protocols spanning ideation to publication, while DeepScientist~\citep{Weng2025DeepScientist} pursues goal-oriented, long-horizon research with explicit objectives to advance state-of-the-art. AI-Researcher~\citep{Tang2025AIResearcher} contributes Scientist-Bench for systematic assessment of autonomous research capabilities. Domain-specific advances continue through AutoLabs~\citep{Panapitiya2025AutoLabs} in chemistry, Curie~\citep{Curie2025} in ML experimentation, and SR-Scientist~\citep{SRScientist2025} in symbolic regression. \emph{Human-AI collaboration} emerges as a critical thrust through frameworks like freephdlabor~\citep{Li2025FreephdLabor}, which architectures research as interactive partnership between human strategists and AI executors, and AI co-scientist~\citep{gottweis2025towards}, which formalizes collaborative research protocols. 
\end{itemize}


  \section{Methodological Integration of AI Scientist Systems}
  \label{sec:method}
  
\begin{figure}[ht]
  \centering
  \includegraphics[width=\linewidth]{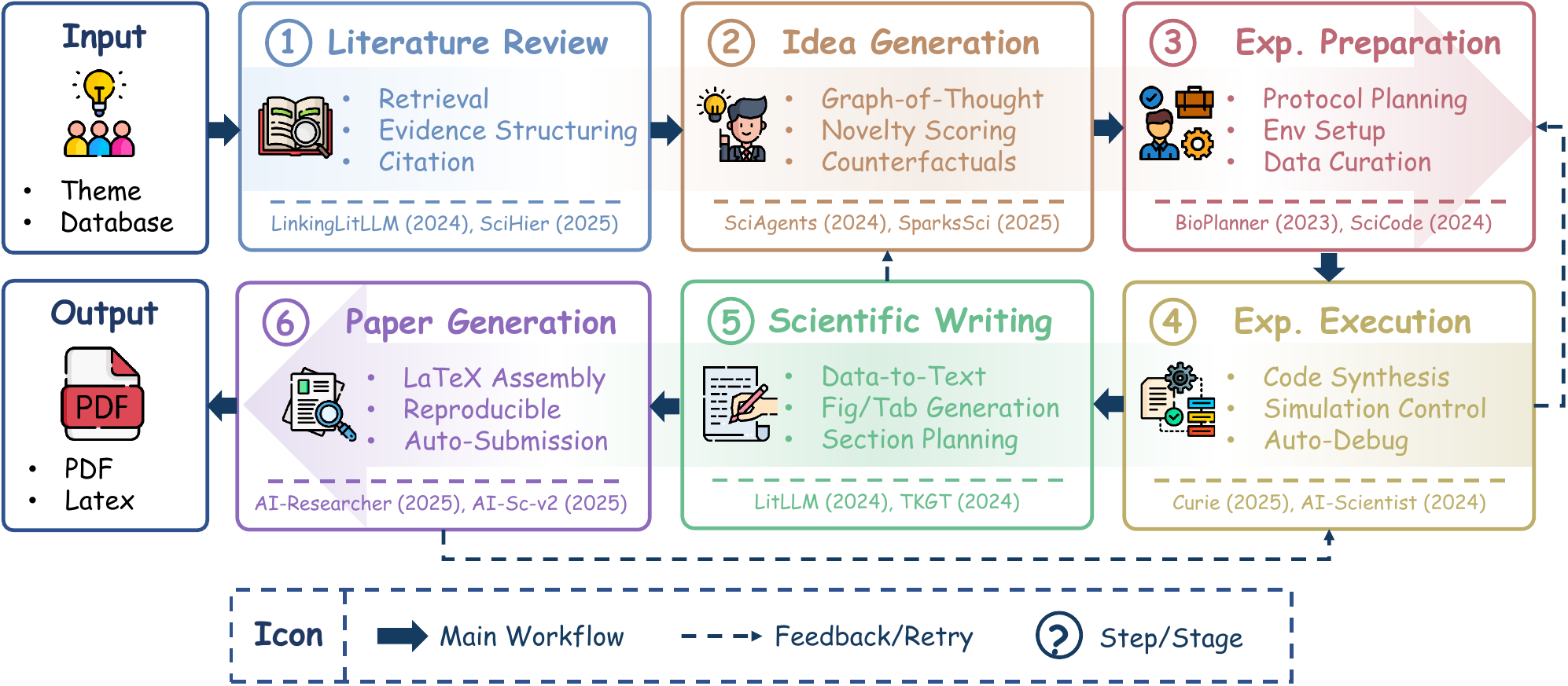}
  \caption{
  \textbf{End-to-end workflow of an AI Scientist system.}
  The six stages represent a closed scientific loop, starting from knowledge synthesis and ending with validated scientific reports.  
  Arrows denote data and reasoning flow, while the outer frame indicates embedded reflection and evaluation mechanisms.
  }
  \label{fig:workflow}
  \end{figure}

  We analyze the methodological components underlying AI Scientist systems through six sequential stages (\textbf{Figure~\ref{fig:workflow}}): \textbf{Literature Review}, \textbf{Idea Generation}, \textbf{Experimental Preparation}, \textbf{Experimental Execution}, \textbf{Scientific Writing}, and \textbf{Paper Generation}, collectively forming a closed-loop autonomous discovery pipeline.

  \subsection{Literature Review}
  \label{sec:lit_review}
  
\begin{figure}[ht]
  \centering
  \includegraphics[width=\linewidth]{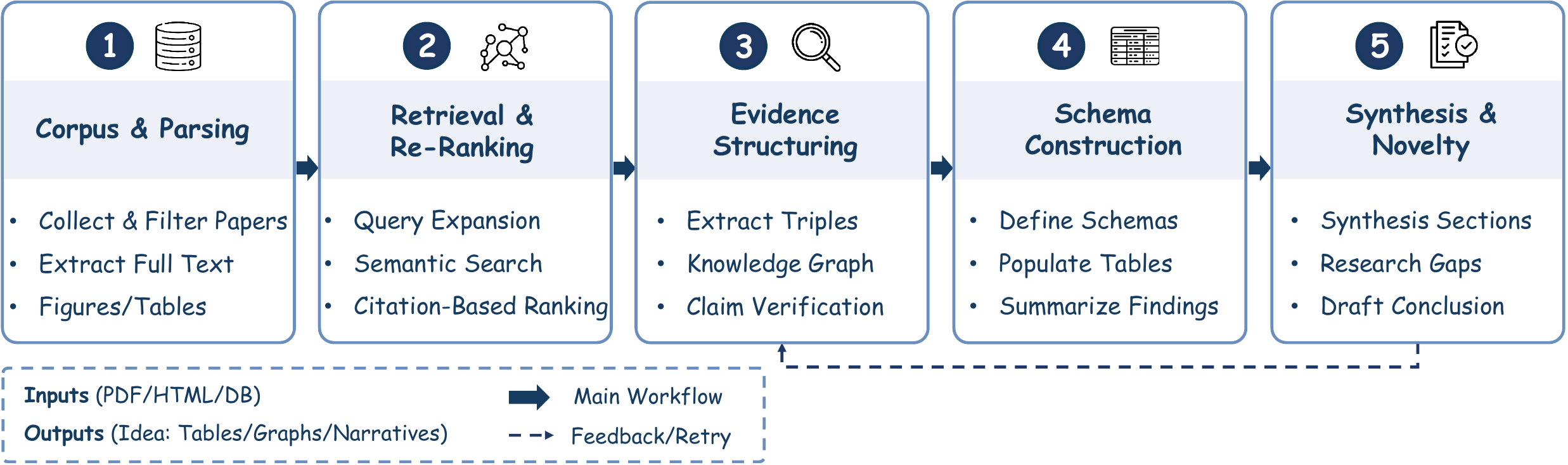}
\caption{\textbf{Pipeline for Automated Literature Review.} 
The workflow transforms unstructured scientific corpora into structured knowledge through five sequential stages with feedback mechanisms enabling iterative refinement.}
  \label{fig:ch3-review}
  \end{figure}

  Literature review transforms unstructured scientific corpora into structured, provenance-aware knowledge representations supporting hypothesis formation. Unlike generic summarization, this stage demands \textbf{scientific faithfulness}, \textbf{verifiable citation grounding}, and \textbf{knowledge-level abstraction}, integrating IR, NLP, and symbolic reasoning into a unified, auditable pipeline (\textbf{Figure~\ref{fig:ch3-review}}).
  
  \begin{itemize}
      \item[\textcolor{hidden-draw}{$\bullet$}] \textbf{\textcolor{hidden-draw}{\underline{Stage~1: Corpus Acquisition and Layout-Aware Parsing.}}} Ingests large-scale scientific repositories (arXiv, PubMed, Semantic Scholar) and transforms heterogeneous document formats into structured representations. Layout-aware parsing~\citep{lo2019s2orc} reconstructs logical document structure—sections, figures, tables, citations—normalizes metadata schemas, and builds citation graphs. This yields indexed corpora preserving textual content and structural semantics essential for downstream reasoning.
  
      \item[\textcolor{hidden-draw}{$\bullet$}] \textbf{\textcolor{hidden-draw}{\underline{Stage~2: Retrieval and Re-Ranking.}}} Optimizes information access through multi-stage ranking pipelines combining complementary strategies. Hybrid approaches leverage sparse lexical matching (BM25) and dense semantic encoders for conceptual similarity. Cross-encoder re-ranking refines candidates using contextualized scoring, while section-aware weighting prioritizes methodology sections. Systems like PaperQA~\citep{Lala2023PaperQA} integrate Retrieval-Augmented Generation, maintaining citation traceability.
  
  \item[\textcolor{hidden-draw}{$\bullet$}] \textbf{\textcolor{hidden-draw}{\underline{Stage~3: Evidence Structuring and Representation Learning.}}} Transforms unstructured textual evidence into machine-interpretable knowledge structures. Information extraction identifies entity mentions and relation tuples, knowledge graph induction maps findings to domain ontologies (UMLS, MeSH), and table synthesis~\citep{Deng2024T3} aggregates quantitative results. All transformations maintain explicit source provenance, enabling verification against original publications.
  
  \item[\textcolor{hidden-draw}{$\bullet$}] \textbf{\textcolor{hidden-draw}{\underline{Stage~4: Schema Induction and Comparative Table Construction.}}} Constructs unified frameworks for cross-study comparison through automated schema discovery. Aspect clustering identifies recurring methodological dimensions (datasets, architectures, metrics), LLM-driven column generation proposes comparison axes, and retrieval-conditioned population extracts values from papers. This aligns heterogeneous studies along standardized dimensions, enabling systematic comparative analysis.
  
  \item[\textcolor{hidden-draw}{$\bullet$}] \textbf{\textcolor{hidden-draw}{\underline{Stage~5: Grounded Narrative Synthesis and Novelty Analysis.}}} Generates coherent research summaries while identifying unexplored directions. Retrieval-conditioned generation produces citation-grounded narratives anchored in source documents, contrastive novelty modeling identifies under-investigated areas, and iterative factuality verification~\citep{Agarwal2024LitLLM} detects inconsistencies. The output includes structured knowledge-gap maps highlighting promising opportunities.
  \end{itemize}

  \subsection{Idea Generation}
  \label{sec:idea_generation}
  
\begin{figure}[ht]
  \centering
  \includegraphics[width=\linewidth]{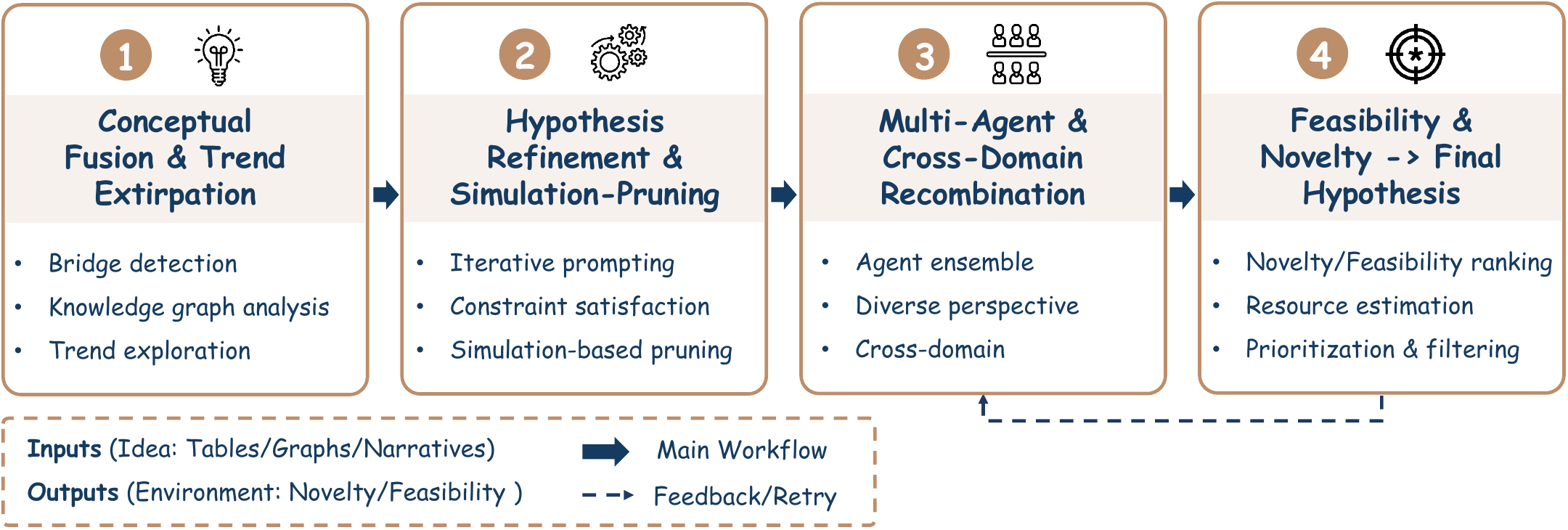}
  \caption{\textbf{Pipeline for Idea Generation.} The workflow transforms structured knowledge into testable hypotheses through four sequential stages supporting creative discovery and feasibility evaluation.}
  \label{fig:ch3-idea}
  \end{figure}

  Idea generation transforms structured knowledge into testable hypotheses through \textbf{scientific creativity}, \textbf{semantic grounding}, and \textbf{novelty control}. We formalize this as a four-stage framework (\textbf{Figure~\ref{fig:ch3-idea}}): conceptual fusion, hypothesis refinement, multi-agent brainstorming, and feasibility evaluation~\citep{Zhou2024HypothesisGenerationLLMs, su2024many}.
  
  \begin{itemize}
      \item[\textcolor{hidden-orange}{$\bullet$}] \textbf{\textcolor{hidden-orange}{\underline{Stage~1: Conceptual Fusion and Trend Extrapolation.}}} Discovers novel research directions by identifying latent connections across disparate knowledge domains. Dense retrieval surfaces conceptually related papers, embedding-based clustering reveals thematic patterns, and knowledge graph reasoning~\citep{Zhou2024HypothesisGenerationLLMs} enables analogical transfer. Temporal citation analysis identifies emerging trends. This constructs novelty-annotated hypothesis banks with estimated originality scores.
  
   \item[\textcolor{hidden-orange}{$\bullet$}] \textbf{\textcolor{hidden-orange}{\underline{Stage~2: Hypothesis Refinement and Knowledge-Grounded Pruning.}}} Filters and refines candidate hypotheses through multi-criteria validation ensuring logical coherence and feasibility. Knowledge graph validation checks consistency with theoretical frameworks, provenance-aware distillation~\citep{vasu2025hyper} traces components to supporting evidence, and simulation-based testing evaluates tractability. Advanced systems employ reinforcement learning for iterative refinement and uncertainty-aware rejection sampling to eliminate speculative hypotheses.
  
   \item[\textcolor{hidden-orange}{$\bullet$}] \textbf{\textcolor{hidden-orange}{\underline{Stage~3: Multi-Agent Brainstorming and Cross-Domain Recombination.}}} Enhances hypothesis diversity through structured collaboration among specialized agents. Multi-role architectures assign distinct responsibilities—critic agents identify flaws, generator agents propose variations, verifier agents check consistency~\citep{su2024many, Ghafarollahi2024SciAgents}. Structured message passing enables agents to ground discussions in retrieved evidence while exploring complementary perspectives, producing hypothesis sets with conceptual diversity.
  
   \item[\textcolor{hidden-orange}{$\bullet$}] \textbf{\textcolor{hidden-orange}{\underline{Stage~4: Feasibility and Novelty Evaluation.}}} Ranks hypotheses through multi-dimensional scoring combining technical feasibility, scientific novelty, and potential impact. Graph-based semantic evaluation quantifies conceptual distance from existing work, while expert-annotated benchmarks~\citep{guo2025ideabench, liu2025researchbench} provide ground-truth assessments. High-scoring hypotheses receive priority routing to experimental preparation, establishing a quality-controlled pipeline from ideation to experimentation.
  \end{itemize}

  \subsection{Experimental Preparation}
  \label{sec:exp_prep}
  
\begin{figure}[ht]
  \centering
  \includegraphics[width=\linewidth]{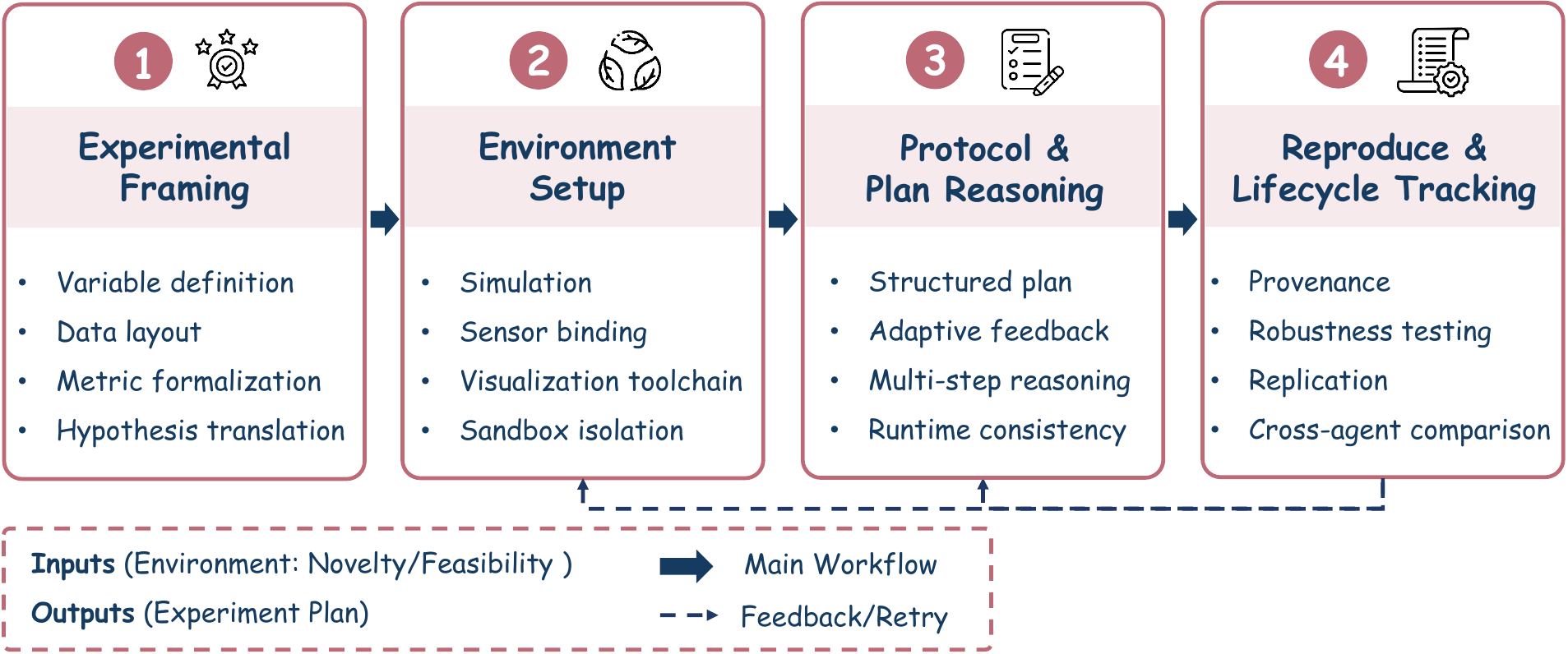}
\caption{\textbf{Pipeline for Experimental Preparation.}  
A four-stage, domain-agnostic workflow that transforms abstract hypotheses into reproducible experiments with integrated provenance tracking and reproducibility guarantees.}
  \label{fig:ch3-exp_prep}
  \end{figure}

  Experimental preparation bridges hypotheses to executable experiments through dataset selection, environment configuration, protocol implementation, and reproducibility tracking. This domain-agnostic, four-stage workflow (\textbf{Figure~\ref{fig:ch3-exp_prep}}) ensures scientific integrity and auditability across physics, chemistry, biology, and data science.
  
  \begin{itemize}
  \item[\textcolor{hidden-red}{$\bullet$}] \textbf{\textcolor{hidden-red}{\underline{Stage 1: Experimental Framing.}}} Translates hypotheses into actionable experimental plans by formalizing variables, metrics, and testable assumptions. Multimodal reasoning over tables and charts~\citep{wu2025tablebench} enables structured variable definition, while automated model template generation~\citep{Majumder2024DiscoveryBench} provides standardized experimental frameworks. This stage establishes clear objectives, measurable outcomes, and explicit constraints for subsequent execution.
  
  \item[\textcolor{hidden-red}{$\bullet$}] \textbf{\textcolor{hidden-red}{\underline{Stage 2: Environment and Instrumentation Setup.}}} Establishes controlled computational and physical environments for reproducible experimentation. Agentic data pipelines configure data backends, structured schema reasoning validates experimental dependencies, and visualization-aware toolchains prepare analysis interfaces. This stage consolidates instrumentation calibration, dependency isolation, and environment sandboxing to ensure experimental validity across computational and laboratory settings.
  
  \item[\textcolor{hidden-red}{$\bullet$}] \textbf{\textcolor{hidden-red}{\underline{Stage 3: Protocol Implementation and Plan Reasoning.}}} Operationalizes experiments through mathematical modeling and multi-step execution planning. Intermediate checkpoints~\citep{guo2024ds} enable continuous introspection and error recovery, while dynamic schema adaptation allows real-time protocol adjustments. This stage transforms high-level experimental designs into executable sequences with built-in monitoring and adaptive re-execution capabilities.
  
  \item[\textcolor{hidden-red}{$\bullet$}] \textbf{\textcolor{hidden-red}{\underline{Stage 4: Reproducibility and Lifecycle Tracking.}}} Maintains comprehensive provenance documentation throughout experimental lifecycles. Parameter tracking, environmental logging~\citep{Gu2024BLADE}, and versioning systems record all experimental configurations and execution details. This stage quantifies cross-agent consistency metrics and ensures verifiable scientific operations through cryptographic hashing, version control, and standardized metadata schemas.
  \end{itemize}

  \subsection{Experimental Execution}
  \label{sec:exp_execution}
  
\begin{figure}[ht]
  \centering
  \includegraphics[width=\linewidth]{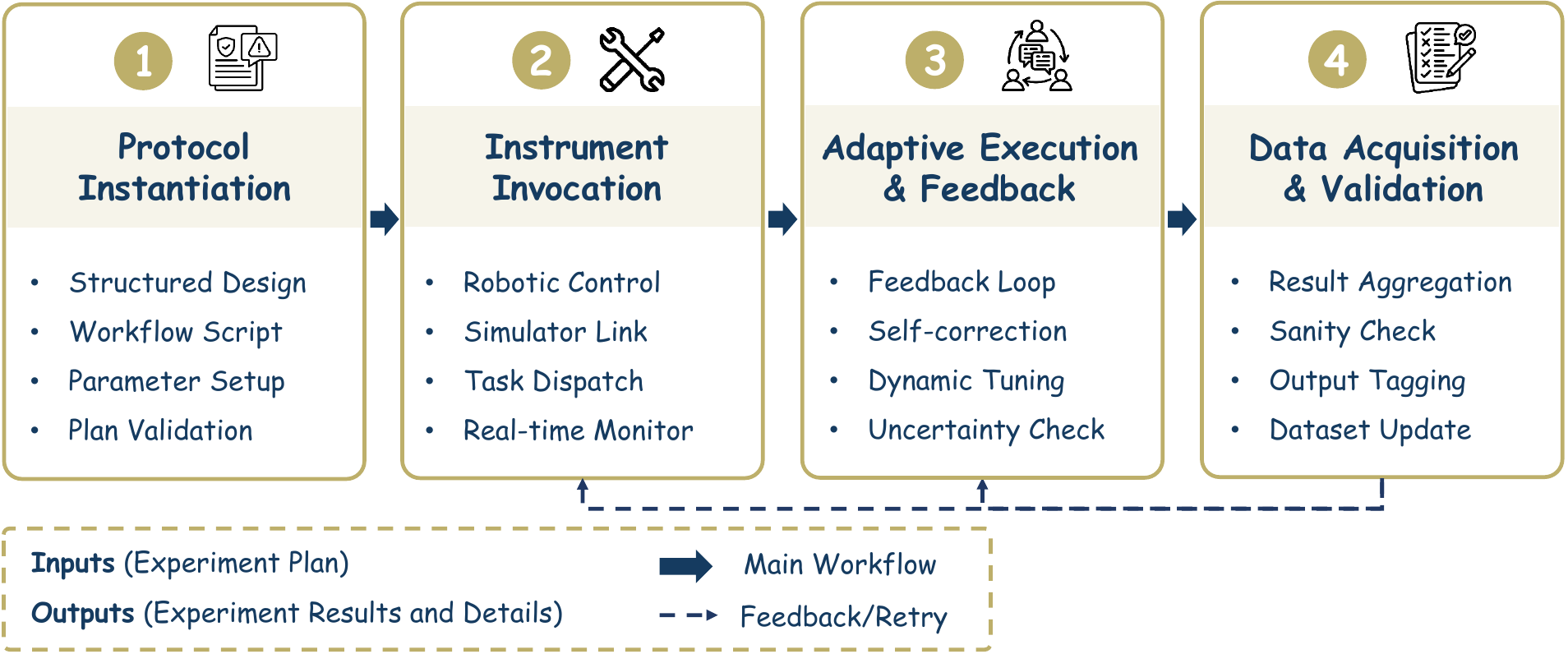}
\caption{\textbf{Pipeline for Experimental Execution.}  
A unified four-stage workflow operationalizing hypothesis-driven protocols into verifiable empirical outcomes through closed-loop reasoning and adaptive control.}
  \label{fig:ch3-exp_exec}
  \end{figure}

  Experimental execution operationalizes hypothesis-driven protocols into verifiable empirical outcomes through closed-loop reasoning and adaptive control~\citep{Boiko2023Coscientist, Curie2025, Kon2025EXPBench, Mandal2025AILA}. Unlike linear automation pipelines, this stage employs continuous feedback between symbolic planning and physical instrumentation, enabling real-time error correction and parameter optimization. We formalize this process through four sequential stages (\textbf{Figure~\ref{fig:ch3-exp_exec}}): Protocol Instantiation, Instrument Invocation, Adaptive Execution, and Data Validation.
  
  \begin{itemize}
  \item[\textcolor{hidden-yellow}{$\bullet$}] \textbf{\textcolor{hidden-yellow}{\underline{Stage 1: Protocol Instantiation.}}}
Translates abstract experimental designs into executable, domain-specific procedures through language-to-action mapping. BioPlanner demonstrates natural-language-to-protocol conversion for biochemical validation~\citep{BioPlanner2023}, while Curie formalizes modular representations enabling task decomposition~\citep{Curie2025}. Hierarchical frameworks encode safety constraints, device mappings, and procedural templates into structured abstraction layers~\citep{Shi2025HierEncapRepProtocol}.
  
  \item[\textcolor{hidden-yellow}{$\bullet$}] \textbf{\textcolor{hidden-yellow}{\underline{Stage 2: Instrument and Tool Invocation.}}}
Interfaces instantiated protocols with heterogeneous experimental infrastructure including robotic laboratories, simulators, and computational clusters. Self-driving laboratories~\citep{Tobias2025SDLReview} integrate robotic manipulation and closed-loop control, while multi-agent systems like ORGANA~\citep{Darvish2025ORGANA} and AutoLabs~\citep{Panapitiya2025AutoLabs} orchestrate apparatus scheduling. Autonomous agents demonstrate scalability through unified API integration and metadata-driven control.
  
  \item[\textcolor{hidden-yellow}{$\bullet$}] \textbf{\textcolor{hidden-yellow}{\underline{Stage 3: Adaptive Execution and Feedback.}}}
Implements real-time monitoring and dynamic protocol revision through closed-loop control. EXP-Bench~\citep{Kon2025EXPBench} and Curie~\citep{Curie2025} leverage intermediate metrics to trigger parameter adjustments. DS-1000~\citep{Lai2023DS1000} and notebook generation~\citep{Yin2022NL2Notebook} demonstrate code-testing feedback loops, while CoScientist~\citep{Boiko2023Coscientist} realizes physical closed-loop reasoning. These frameworks converge reinforcement learning, self-critique, and error detection into adaptive control.
  
  \item[\textcolor{hidden-yellow}{$\bullet$}] \textbf{\textcolor{hidden-yellow}{\underline{Stage 4: Data Acquisition and Validation.}}}
Consolidates multimodal experimental outputs into provenance-linked, validated datasets through comprehensive metadata annotation and statistical verification~\citep{Mandal2025AILA}. Curie~\citep{Curie2025} and EXP-Bench~\citep{Kon2025EXPBench} embed runtime analytics validating outputs against simulations. Physics applications aggregate sensor telemetry for causal interpretation. This establishes an auditable evidence chain linking execution logs to scientific reporting.
  \end{itemize}

  \subsection{Scientific Writing}
  \label{sec:sci_writing}
  
\begin{figure}[ht]
      \centering
      \includegraphics[width=1\linewidth]{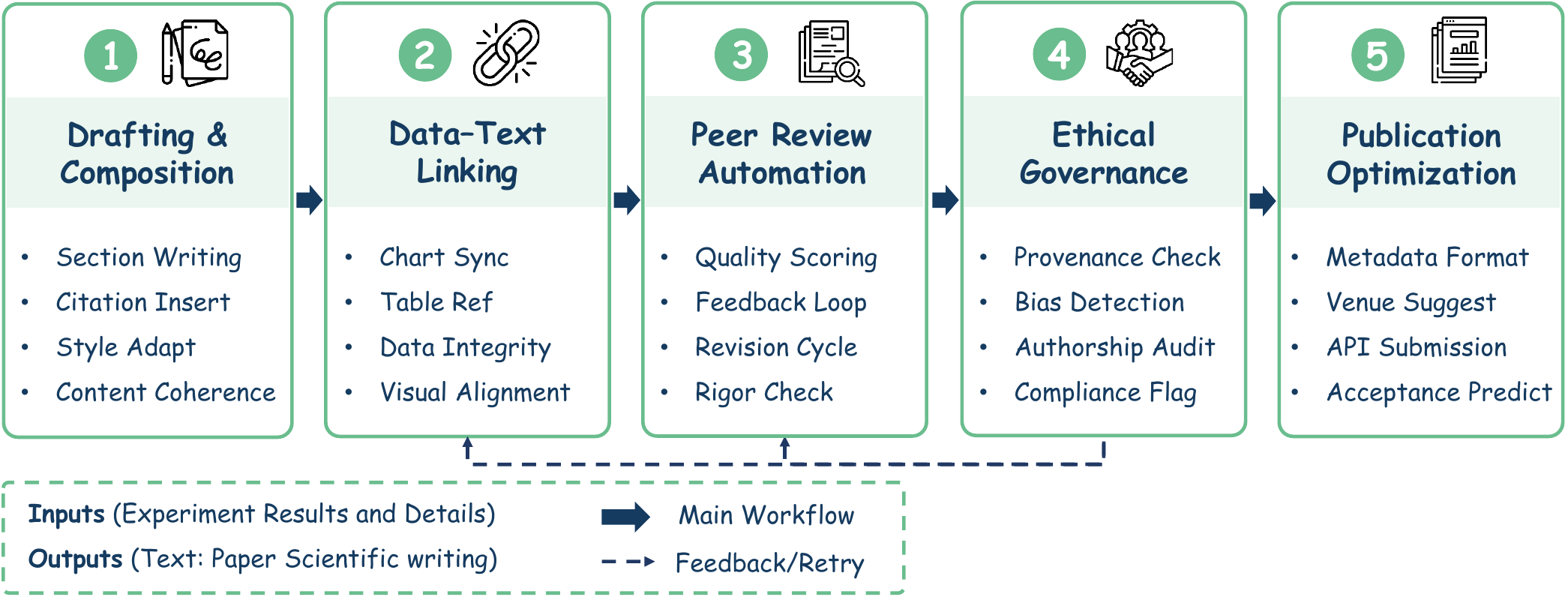}
\caption{
\textbf{Pipeline for Scientific Writing.}
A five-stage workflow transforming experimental findings into verifiable, publication-grade scholarly narratives through multimodal composition and ethical governance.
}
      \label{fig:ch3-sci_write}
  \end{figure}

  Scientific writing transforms experimental findings into verifiable, publication-grade scholarly narratives through factual grounding, data–text alignment, and ethical compliance. LLM-assisted writing now permeates biomedical publishing, educational research~\citep{li2024exploring}, and simulation science~\citep{cheng2025artificial}. We formalize this ecosystem through five interlinked stages (\textbf{Figure~\ref{fig:ch3-sci_write}}): Drafting, Data–Text Linking, Peer-Review Automation, Ethical Governance, and Publication Optimization.
  
  \begin{itemize}
  \item[\textcolor{hidden-green}{$\bullet$}] \textbf{\textcolor{hidden-green}{\underline{Stage~1: Drafting and Structural Composition.}}}
Converts analytical outputs into disciplinary-compliant academic prose through logical structuring and cross-sectional coherence~\citep{telenti2024large, khalifa2024using}. Domain-specific assistants achieve expert-level fluency via section-specific prompts~\citep{cheng2025artificial}, while citation-aware mechanisms map claims to verified sources and structure-aware templates enforce parallelism. Human-AI co-editing workflows reduce grammatical noise and accelerate drafting~\citep{guo2024artificial}.
  
  \item[\textcolor{hidden-green}{$\bullet$}] \textbf{\textcolor{hidden-green}{\underline{Stage~2: Data–Text Linking and Multimodal Representation.}}}
Aligns quantitative data with narratives through automated cross-referencing of tables and figures~\citep{ WritingBench2025Benchmark}. Hybrid language-vision models generate visual elements from structured data~\citep{salvagno2023can}, while grounded captioning mechanisms automate caption rewriting and figure suggestion~\citep{son2025ai}. This establishes scientific writing as multimodal composition where LLMs mediate between data, visuals, and text.
  
  \item[\textcolor{hidden-green}{$\bullet$}] \textbf{\textcolor{hidden-green}{\underline{Stage~3: Peer-Review Automation and Quality Verification.}}}
Automates manuscript evaluation through LLM-based reviewers trained on editorial corpora, identifying inconsistent claims and logical fallacies~\citep{zhou2025large, son2025ai}. Systems emulate referee workflows via structured comments and quantitative scores, implementing iterative review loops that reduce latency~\citep{feyzollahi2025adoption}. Reproducibility and algorithmic bias remain challenges, necessitating transparent confidence calibration~\citep{telenti2024large}.
  
  \item[\textcolor{hidden-green}{$\bullet$}] \textbf{\textcolor{hidden-green}{\underline{Stage~4: Ethical Compliance and Authorship Governance.}}}
Enforces mandatory disclosure of AI involvement to prevent ghostwriting~\citep{ateriya2025exploring, resnik2025ethics, Mwita2025AIWritingEthics}. Standardized frameworks document prompt histories, model identifiers, and revision provenance~\citep{roe2023review}, while layered attribution models recognize human intellectual contribution and algorithmic assistance~\citep{jo2023promise}. These mechanisms anchor trust and accountability in AI-augmented authorship.
  
  \item[\textcolor{hidden-green}{$\bullet$}] \textbf{\textcolor{hidden-green}{\underline{Stage~5: End-to-End Publication Optimization.}}}
Integrates drafting, multimodal generation, review, and ethics into unified workflows. Frameworks like WritingBench~\citep{WritingBench2025Benchmark} and SPOT~\citep{son2025ai} evaluate manuscripts for factual accuracy and citation soundness. Adaptive agents monitor document lifecycles, suggesting figure adjustments, formatting references, and validating ethical disclosures, enabling scientists to transition to supervisory orchestration~\citep{khalifa2024using}.
  \end{itemize}

  \subsection{Paper Generation}
  \label{sec:paper_generation}
  
\begin{figure}[ht]
    \centering
    \includegraphics[width=0.9\linewidth]{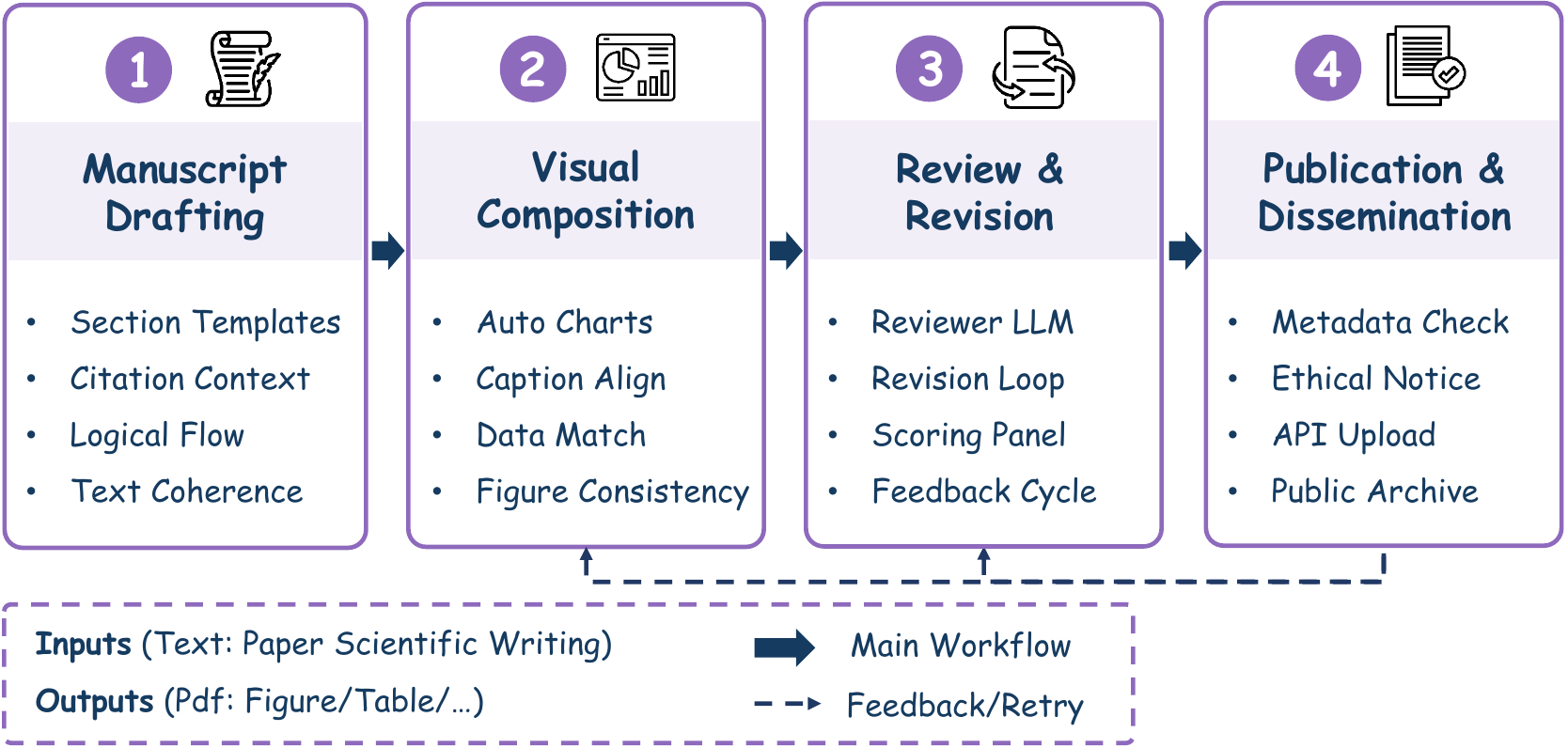}
\caption{
  \textbf{Pipeline for Paper Generation.} A unified four-stage workflow integrating ideation, experimentation, visualization, composition, and review into autonomous manuscript creation.
}
    \label{fig:ch3-paper_gen}
  \end{figure}

  Paper generation integrates ideation, experimentation, visualization, composition, and review into autonomous manuscript creation workflows (\textbf{Figure~\ref{fig:ch3-paper_gen}}). Systems like The AI Scientist v1/v2 generate complete papers—including figures, experiments, and reviews—with minimal human oversight~\citep{Lu2024AIScientist, Yamada2025AIScientistV2}, while AI-Researcher establishes end-to-end pipelines and benchmarks (Scientist-Bench) for systematic evaluation~\citep{Tang2025AIResearcher}. We decompose this process into four stages: Manuscript Drafting, Visual \& Tabular Composition, Review \& Revision, and Publication Dissemination.
  
  \begin{itemize}
  \item[\textcolor{hidden-purple}{$\bullet$}] \textbf{\textcolor{hidden-purple}{\underline{Stage 1: Manuscript Drafting.}}}
Transforms structured outputs into full manuscript sections through retrieval-augmented generation. The AI Scientist v1 converts experiment logs to LaTeX drafts, synthesizes literature context, and embeds citations automatically~\citep{Lu2024AIScientist}. AIGS extends this via multi-agent pipelines for full-cycle generation~\citep{Liu2024AIGS}, employing template-based formatting for journal compliance.
  
  \item[\textcolor{hidden-purple}{$\bullet$}] \textbf{\textcolor{hidden-purple}{\underline{Stage 2: Visual \& Tabular Composition.}}}
Generates and integrates visual artifacts with narrative alignment. The AI Scientist converts experimental outputs to plots and tables, generating captions and inserting cross-references~\citep{Lu2024AIScientist}, while SciSciGPT integrates data processing with figure generation~\citep{Shao2025SciSciGPT}. Core technologies include automatic plotting, figure-caption co-generation, and multimodal consistency validation.
  
  \item[\textcolor{hidden-purple}{$\bullet$}] \textbf{\textcolor{hidden-purple}{\underline{Stage 3: Review \& Revision Agent.}}}
Embeds internal reviewer agents to critique and refine manuscripts. The AI Scientist v2 implements agentic tree-search with dedicated reviewer components generating structured feedback~\citep{Yamada2025AIScientistV2}, while AI-Researcher enables automated revision loops~\citep{Tang2025AIResearcher}. Key methodologies include peer-review-trained models, LLM-driven comment generation, and iterative author-reviewer simulation.
  
  \item[\textcolor{hidden-purple}{$\bullet$}] \textbf{\textcolor{hidden-purple}{\underline{Stage 4: Publication Dissemination.}}}
Prepares manuscripts for submission through venue-specific formatting and metadata management. Automated pipelines ensure submission readiness~\citep{madanchian2024ai}, while post-publication modules enable versioning and citation tracking. Core technologies include template-mapping tools, submission-API integration, metadata provenance logging, and ethical-disclosure verification.
  \end{itemize}


  \section{Applications of AI Scientist Systems}
  \label{sec:applications}
  
  AI Scientist systems now autonomously generate hypotheses, design experiments, analyze data, and produce manuscripts across diverse scientific domains. We categorize existing implementations into two tiers: (1)~\textbf{General AI Scientist Systems}, pursuing end-to-end, cross-domain autonomy; and (2)~\textbf{Domain-Specific AI Scientist Systems}, specializing in chemistry, biology, physics, and meta-science.
  
  
  \subsection{General AI Scientist Systems}
  \label{sec:general_ai_scientist}
  
  General AI Scientist systems embody domain-agnostic frameworks replicating the complete scientific workflow—from question formulation to experiment execution, result interpretation, and dissemination—serving as both discovery accelerators and epistemological probes into machine intelligence.

  \begin{itemize}[leftmargin=1.2em]
       \item[\textcolor{hidden-draw}{$\bullet$}] \textbf{\textcolor{hidden-draw}{The AI Scientist v1}}~\citep{Lu2024AIScientist} deploys modular multi-agent architecture—planning, coding, analysis, writing—coordinated by a meta-scientist module, autonomously selecting topics, generating executable code, drafting papers, and reproducing canonical machine learning studies.
  
      \item[\textcolor{hidden-draw}{$\bullet$}] \textbf{\textcolor{hidden-draw}{The AI Scientist v2}}~\citep{Yamada2025AIScientistV2} advances v1 through agentic tree-search planning, dynamically exploring parallel hypotheses and evaluating novelty/validity via reflective feedback loops.
  
       \item[\textcolor{hidden-draw}{$\bullet$}] \textbf{\textcolor{hidden-draw}{AI-Researcher}}~\citep{Tang2025AIResearcher} prioritizes transparency through provenance-tracking memory graphs recording all artifacts—code, data logs—co-developed with Scientist-Bench for reproducibility evaluation.
  
      \item[\textcolor{hidden-draw}{$\bullet$}] \textbf{\textcolor{hidden-draw}{Curie}}~\citep{Curie2025} achieves rigorous experimental control via causality-aware planning loops, automating machine learning hypothesis testing with explicit causal assumptions.
  \end{itemize}
  
  \subsection{Chemistry and Materials Science}
  \label{sec:chemistry}
  
  Chemistry and materials science provide mature testbeds for AI Scientist systems through structured molecular representations, well-defined protocols, and self-driving laboratories (SDLs), enabling closed-loop integration of digital reasoning and physical experimentation for \textit{de novo} discovery.
  
  \begin{itemize}[leftmargin=1.2em]
      \item[\textcolor{hidden-orange}{$\bullet$}]  \textbf{\textcolor{hidden-orange}{Coscientist}}~\citep{Boiko2023Coscientist} integrates GPT-4 reasoning with robotic liquid handlers, autonomously planning reactions, generating control code, interpreting spectroscopic feedback, and iteratively refining hypotheses.
  
      \item[\textcolor{hidden-orange}{$\bullet$}]  \textbf{\textcolor{hidden-orange}{A-Lab}}~\citep{Szymanski2023AutonomousLab} combines Bayesian optimization with LLM-guided experiment design, autonomously synthesizing and characterizing thousands of novel inorganic materials weekly.
      
      \item[\textcolor{hidden-orange}{$\bullet$}]  \textbf{\textcolor{hidden-orange}{Robotic AI Chemist}}~\citep{Song2024RoboticAIchemist} converges cognitive autonomy (literature-grounded reasoning) with physical autonomy (robotic manipulation) for end-to-end reaction design and execution.
  
      \item[\textcolor{hidden-orange}{$\bullet$}]  \textbf{\textcolor{hidden-orange}{AutoLabs}}~\citep{Panapitiya2025AutoLabs} deploys multi-agent self-correction cycles—planning, control, safety auditing—detecting anomalies and recalibrating instruments for enhanced throughput and safety.
  \end{itemize}
  
  \subsection{Biology and Biomedical Research}
  \label{sec:biology}
  
  Biology and biomedicine demand semantic interpretation of complex protocols and causal inference from noisy, high-dimensional data, targeting disease pathway discovery and drug target identification.
  
  \begin{itemize}[leftmargin=1.2em]
      \item[\textcolor{hidden-red}{$\bullet$}]  \textbf{\textcolor{hidden-red}{BioPlanner}}~\citep{BioPlanner2023} establishes benchmarks for LLM-driven biological protocol design, formalizing translation of research goals into executable experimental workflows.
  
      \item[\textcolor{hidden-red}{$\bullet$}]  \textbf{\textcolor{hidden-red}{LLM4GRN}}~\citep{Afonja2024LLM4GRN} integrates LLM reasoning with bioinformatics tools to discover causal gene regulatory networks, automating complex data analysis for biological mechanism inference.
  
      \item[\textcolor{hidden-red}{$\bullet$}]  \textbf{\textcolor{hidden-red}{Hierarchically Encapsulated Representation}}~\citep{Shi2025HierEncapRepProtocol} employs hierarchical architectures for multi-level protocol reasoning—from macro workflows to micro parameters—enabling robust, context-aware self-driving lab planning.
  \end{itemize}

  \subsection{Physics and Engineering}
  \label{sec:physics}
  
  Physics and engineering pursue fundamental equation discovery through abductive inference—extracting symbolic principles from observational data via synergistic integration of numerical simulation, symbolic reasoning, and real-time instrumentation control.
  
  \begin{itemize}[leftmargin=1.2em]
      \item[\textcolor{hidden-green}{$\bullet$}] \textbf{\textcolor{hidden-green}{Agentic Physics Experiments}} deploys AI agents at particle accelerator facilities, autonomously coordinating data acquisition and beamline configuration via closed feedback loops for optimized calibration.
  
      \item[\textcolor{hidden-green}{$\bullet$}] \textbf{\textcolor{hidden-green}{SR-Scientist}}~\citep{SRScientist2025} employs agentic workflows for symbolic regression, autonomously discovering fundamental equations from observational data.
  
      \item[\textcolor{hidden-green}{$\bullet$}] \textbf{\textcolor{hidden-green}{AI Feynman}}~\citep{Udrescu2020AIPhysRev} introduces physics-constrained symbolic search (dimensional consistency), rediscovering classical laws from raw data and inspiring modern symbolic reasoning architectures.
  
      \item[\textcolor{hidden-green}{$\bullet$}] \textbf{\textcolor{hidden-green}{Quantum-Agent-SDL}}~\citep{Cao2024QuantumAgentSDL} combines reinforcement learning with LLM-based hypothesis refinement for self-optimized qubit calibration and error correction in quantum computing.
  \end{itemize}
  
  \subsection{Meta-Science and Social Science}
  \label{sec:metascience}
  
  Meta-science applications analyze the scientific enterprise itself—mapping knowledge flows, identifying paradigms, assessing reproducibility—transitioning AI systems from research executors to meta-researchers studying science as a complex adaptive system.
  
  \begin{itemize}[leftmargin=1.2em]
      \item[\textcolor{hidden-purple}{$\bullet$}] \textbf{\textcolor{hidden-purple}{SciAgents}}~\citep{Ghafarollahi2024SciAgents} employs multi-agent collaboration and dynamic graph reasoning to traverse publication/author networks, identifying knowledge gaps, interdisciplinary connections, and predicting research trends.
  
      \item[\textcolor{hidden-purple}{$\bullet$}] \textbf{\textcolor{hidden-purple}{AI for Social Science (AI4SS)}}~\citep{xu2024ai} outlines roadmaps for AI-driven large-scale social modeling, policy simulation, and analysis of innovation diffusion and collaboration patterns.
  
      \item[\textcolor{hidden-purple}{$\bullet$}] \textbf{\textcolor{hidden-purple}{Ethical Governance Frameworks}}~\citep{resnik2025ethics, Mwita2025AIWritingEthics} address societal implications of AI-generated science, proposing frameworks for authorship management, credit attribution, and accountability maintenance.
  \end{itemize}
\section{Open Problems and Future Directions}
\label{sec:problem}
The AI Scientist paradigm has achieved substantial methodological progress, yet fundamental challenges remain in transitioning from domain-specific demonstrations to general-purpose, verifiable research automation. Current systems exhibit critical limitations in reproducibility guarantees, uncertainty quantification, cross-domain generalization, and ethical governance. Based on our systematic analysis, we identify four interdependent research frontiers essential for advancing AI Scientists from experimental prototypes to trustworthy scientific instruments.

\begin{itemize}
\item \textbf{Reproducibility-by-Design and Verifiable Science.}
Multi-stage autonomous systems amplify reproducibility challenges through compositional complexity, where parameter variations in early stages cascade into divergent outcomes. Achieving verifiable science requires architectural integration of three components: \emph{(1)~Environmental Determinism}—containerization of computational dependencies, cryptographic hashing of data artifacts, and version control of model checkpoints ensure bitwise reproducibility; \emph{(2)~Fine-Grained Provenance}—cryptographic linking of manuscript claims to specific code commits, data versions, and model parameters enables claim-level verification~\citep{Gu2024BLADE}; and \emph{(3)~Automated Verification}—AI-driven auditor agents performing logical consistency checks, detecting claim-evidence misalignments, and validating statistical inferences before publication~\citep{son2025ai}. These mechanisms must transition from post-hoc validation to intrinsic architectural properties.

\item \textbf{Uncertainty Quantification and Epistemic Humility.}
Current AI Scientists exhibit overconfidence, masking underlying uncertainties and alternative explanatory hypotheses. Principled uncertainty quantification requires architectures that explicitly model epistemic uncertainty (knowledge limitations) and aleatoric uncertainty (inherent stochasticity) throughout the research workflow. Promising approaches include Bayesian neural networks for confidence-calibrated predictions, ensemble methods maintaining diverse hypothesis sets~\citep{Yamada2025AIScientistV2}, and conformal prediction providing distribution-free uncertainty intervals. Beyond technical mechanisms, systems must demonstrate epistemic humility—recognizing knowledge boundaries, expressing calibrated confidence, requesting additional experimentation when evidence is insufficient, and deferring to domain experts for decisions requiring contextual judgment. This shift from deterministic outputs to probabilistic reasoning with explicit uncertainty bounds is essential for trustworthy scientific automation.

\item \textbf{Cross-Domain Generalization through Modular Composition.}
Current AI Scientists achieve domain-specific success~\citep{Boiko2023Coscientist} but exhibit limited transferability to fields with informal methodologies or rapidly evolving instrumentation. Monolithic end-to-end architectures encode domain assumptions implicitly, preventing compositional reuse of learned capabilities. Cross-domain generalization requires transition toward modular, composable architectures based on capability factorization. This paradigm decomposes autonomous research into reusable functional modules—causal inference, symbolic regression, statistical testing, data visualization—each with formally specified input-output contracts. A meta-planning layer dynamically composes these modules into domain-adapted workflows, enabling procedural transfer through novel module assemblies rather than domain-specific retraining. Standardization of module interfaces, analogous to software engineering's API contracts, would facilitate community-driven development of specialized capabilities while preserving interoperability. This architectural shift from monolithic systems to composable toolkits represents a prerequisite for general-purpose scientific automation.

\item \textbf{Human-AI Collaboration and Ethical Governance.}
The frontier of impactful research automation lies not in full autonomy but in synergistic human-AI partnerships that leverage complementary strengths—human creativity, intuition, and contextual judgment combined with AI's scalability, systematic exploration, and computational power. Frameworks exemplified by freephdlabor~\citep{Li2025FreephdLabor} demonstrate interactive research automation where human strategists guide exploration while AI executors perform systematic experimentation and analysis. Realizing this vision requires formal protocols for role allocation, mixed-initiative planning, and explanation generation enabling meaningful human oversight. Concurrently, ethical governance frameworks become imperative as AI-generated research enters the scientific record. Essential components include machine-readable author contribution statements distinguishing human intellectual input from AI execution~\citep{resnik2025ethics}, cryptographically verifiable audit trails documenting decision provenance, and risk-gated execution mechanisms requiring human approval for high-stakes domains~\citep{ateriya2025exploring}. Establishing community-wide standards for AI authorship attribution, intellectual property rights, and accountability mechanisms represents both a technical challenge—requiring standardized metadata schemas and verification protocols—and a societal imperative for maintaining scientific integrity and public trust in AI-augmented research.
\end{itemize}

\section{Conclusion}\label{sec:conclusion}

This survey charts the AI Scientist's evolution from fragmented specialized tools to integrated, end-to-end research agents through a unified six-stage methodological framework and three-phase historical narrative. Our synthesis reveals increasing integration, self-reflection, and autonomy. The current frontier pursues dual objectives: scalability and impact through systems like The AI Scientist, and sophisticated human-AI collaboration through interactive frameworks. While challenges in generalizability, robustness, and ethical governance remain formidable, the AI Scientist's promise lies not in replacing human researchers, but in forging symbiotic partnerships that augment human creativity with machine-scale exploration, fundamentally redefining scientific inquiry and accelerating discovery.


\bibliographystyle{unsrt}
\bibliography{sample-base}


\end{document}